\documentclass[runningheads]{llncs}
\usepackage[T1]{fontenc}
\usepackage{multirow}
\usepackage{graphicx}
\usepackage{amsmath}
\usepackage{amssymb}
\usepackage{booktabs}

\usepackage{xcolor}
\begin{document}
\title{SDMixer: Sparse Dual-Mixer for Time Series Forecasting}
\author{Xiang Ao\inst{1}\orcidID{0009-0001-3215-8334} }
\authorrunning{X. Ao}
\institute{School of Software Engineering, Beijing Jiaotong University, Beijing, China \thanks{A version of this work was peer-reviewed and accepted by the DSFA track of PAKDD 2026. The authors withdrew the paper from the formal proceedings due to a lack of institutional funding.}
\email{ao.xiang.axel@outlook.com} or \email{24301058@bjtu.edu.cn}}

\maketitle              % typeset the header of the contribution
\begin{abstract}
Multivariate time series forecasting is widely applied in fields such as transportation, energy, and finance. However, the data commonly suffers from issues of multi-scale characteristics, weak correlations, and noise interference, which limit the predictive performance of existing models. This paper proposes a dual-stream sparse Mixer prediction framework that extracts global trends and local dynamic features from sequences in both the frequency and time domains, respectively. It employs a sparsity mechanism to filter out invalid information, thereby enhancing the accuracy of cross-variable dependency modeling. Experimental results demonstrate that this method achieves leading performance on multiple real-world scenario datasets, validating its effectiveness and generality. The code is available at:\url{https://github.com/SDMixer/SDMixer}.
\end{abstract}
\section{Introduction}

Multivariate time series forecasting holds significant application value in fields such as energy scheduling, traffic management, financial forecasting, and industrial monitoring \cite {Energy}\cite{Traffic}\cite{Financial} \cite{Industrial}. However, real-world time series are often composed of components with various frequencies and amplitudes, encompassing weakly correlated features, noise perturbations, and complex temporal dependencies. These composite structural characteristics pose two core challenges for models: First, in high-dimensional variable spaces, dependencies exhibit pronounced sparsity and heterogeneity, with different variables contributing significantly differently to future outcomes. Spurious correlations frequently mislead model learning. Second, temporal and frequency domain features exhibit learning biases during unified modeling, causing models to favor patterns with larger amplitudes and lower entropy, thereby overlooking critical yet subtle information changes, particularly evident in long-term forecasting tasks \cite{Autoformer}.

In recent years, time series forecasting models have witnessed rapid development. Transformer-based models \cite{Attention} are capable of capturing global dependencies and excel in long-sequence forecasting. For instance, Informer \cite{Informer} reduces computational costs through probabilistic sparse attention, while iTransformer \cite{iTransformer} optimizes attention mechanisms across variable dimensions. However, these models suffer from high computational complexity and are susceptible to noise interference in multivariate scenarios \cite{TSMixer}. To mitigate their degradation in long-term forecasting, some studies have approached modeling from the frequency domain. For example, Autoformer \cite{Autoformer} introduces an autocorrelation mechanism for spectral decomposition, FEDformer \cite{FEDformer} utilizes Fourier transforms to model multi-scale variations, and TimesNet \cite{TimesNet} extracts multi-frequency components through temporal convolutions, significantly enhancing trend and weak signal capture capabilities. Nevertheless, these methods often lack explicit modeling of the sparsity of variable dependencies, allowing redundant variables to still propagate noise interference during modeling. Additionally, the fusion of temporal and frequency domain features mostly remains at the stage of late concatenation, insufficiently leveraging their complementary nature during the modeling process \cite{Crossformer}. Therefore, how to effectively decouple temporal-frequency structures while ensuring model efficiency and stability, and enhancing attention to key variables and weak signals, remains an area requiring further in-depth research.

To this end, this paper proposes a dual-stream sparse Mixer prediction framework, SDMixer (Sparse Dual-Mixer for Time Series Forecasting). This method centers on the decoupling of temporal and frequency domain information. In the frequency domain stream, it explicitly enhances the representation capabilities of different frequency signal components to overcome the degradation of weak signals during modeling. Simultaneously, in the temporal domain stream, it introduces a learnable sparse selection mechanism to adaptively filter out ineffective variable dependencies and reduce noise interference. Furthermore, SDMixer employs lightweight feature mixers instead of complex global attention structures, maintaining cross-variable and temporal dimension interactions while effectively controlling computational complexity and parameter count, enhancing training stability and engineering deployment efficiency. Ultimately, the fusion of the dual streams achieves complementary temporal-frequency information, enabling the model to accurately capture trend changes and fine-grained dynamics of sequences in complex contexts. Experimental results demonstrate that SDMixer significantly outperforms mainstream baseline models in terms of forecasting performance across multiple real-world datasets. The contributions of this study can be summarized as follows:

\begin{enumerate}
    \item Proposing the dual-stream sparse Mixer structure, SDMixer, which combines temporal-frequency decoupled modeling with sparse dependency filtering to efficiently capture key variables and weak frequency components, significantly improving model robustness.
    \item Utilizing lightweight feature mixers instead of attention structures to reduce computational complexity and inference costs, enabling the model to possess better potential for engineering deployment and large-scale adaptability.
    \item Achieving superior forecasting performance compared to advanced methods on multiple real-world datasets, validating the effectiveness of our approach.

\end{enumerate}

\section{Related Work}
Early time series forecasting methods were based on statistics or traditional machine learning \cite{Time_series_analysis}\cite{ARIMA+}. In recent years, deep learning methods have developed rapidly, evolving from traditional RNN-based models (e.g., LSTM, GRU) to convolutional models \cite{LSTM}\cite{ST-GCN} and attention-based models \cite{Autoformer}\cite{FEDformer}, which have effectively improved long-term forecasting performance. However, these models generally assume dense dependencies between variables, making them incapable of handling sparse interaction structures in real-world data and prone to generating spurious correlations \cite{CSDI}.

To alleviate the degradation of long-term dependencies, frequency-domain and structured modeling approaches have gained attention. For instance, frequency modeling based on wavelets and FFT \cite{WPMixer}\cite{Autoformer}, as well as time-frequency hybrid forecasting networks \cite{TimesNet}\cite{Crossformer}, have all demonstrated that the frequency domain exhibits stronger representational power for long-term trends. Meanwhile, techniques such as sparse attention and graph-structured modeling have been employed to enhance focus on key variables \cite{Graph_WaveNet}\cite{DST-GCN}\cite{DGNN}, but graph structure construction relies on prior knowledge and is difficult to adapt to dynamic changes.

In recent years, lightweight structures based on the MLP-Mixer architecture have shown great potential in computer vision and time series domains \cite{MLP-Mixer}\cite{SCINet}. These structures utilize axial feature mixing to replace global attention, significantly reducing computational costs. Nevertheless, existing methods typically perform single-stream modeling only in the temporal or feature dimension, lacking the ability to jointly model frequency patterns and sparse dependencies. As a result, they struggle to adapt to issues such as time-varying dominant frequencies in composite signals and the easy neglect of weak signals.

To address these limitations, this paper proposes the SDMixer, which synergistically improves the performance of multivariate time series forecasting through time-frequency decoupling and sparse dependency filtering.

\section{Problem Description}

Multivariate Time Series Forecasting (MTSF) aims to utilize historical observations of multiple interrelated variables evolving over time to infer future trends and variation patterns. Suppose there exists a sequence consisting of \( C \) variables, which are observed simultaneously at discrete time points \( t \). The historical observations over the time window \([t-L+1, t]\) are denoted as:

\begin{equation}
X = [x_{t-L+1}, x_{t-L+2}, \ldots, x_{t}] \in \mathbb{R}^{L \times C},
\end{equation}
where \( L \) represents the length of the input historical sequence, \( C \) is the number of variables, and \( x_{\tau} \in \mathbb{R}^{C} \) denotes the multivariate observation vector at time \( \tau \). In the forecasting task, the goal is to learn a functional mapping \( f_{\theta}(\cdot) \) based on \( X \) that can predict the sequence for the next \( L' \) time steps:

\begin{equation}
\hat{Y} = f_{\theta}(X) \in \mathbb{R}^{L' \times C},
\end{equation}
where \( L' \) is the prediction length, \( \hat{Y} = [\hat{x}_{t+1}, \ldots, \hat{x}_{t+L'}] \) is the prediction result, and \( \hat{x}_{\tau} \in \mathbb{R}^{C} \).

\section{Method}

\subsection{Overview of SDMixer}

Real-world multivariate time series often contain both smoothly evolving trend components and seasonal components characterized by periodic oscillations. These two types of components exhibit significant differences in their spectral energy distribution. Trends typically correspond to low-frequency, high-energy patterns, while periodic structures may possess lower energy but carry crucial predictive information. Therefore, modeling solely in the time domain can easily result in weak periodic signals being overshadowed by trend components, biasing the model towards prominent patterns.

To address this issue, SDMixer first maps the input sequence to the frequency domain and performs structured decomposition based on the criterion of energy dominance. Let the input sequence be \(X \in \mathbb{R}^{B \times L \times C},\)where \( B \) is the batch size, \( L \) is the historical window length, and \( C \) is the number of variables. We employ the Fast Fourier Transform (FFT) to extract the frequency spectrum as \(X_F = \text{FFT}(X)\), where \( X_F \) is the complex frequency-domain representation, containing both phase and energy information. Since the energy of different frequency components reflects their contribution to the overall structure, we compute their magnitude spectrum as \(A = |X_F|\).

A larger magnitude indicates that the corresponding frequency mode has a stronger dominant role in the evolution of the time series. To preserve the main structural information, we select the top-K frequencies with the highest energy in each variable channel:
\begin{equation}
    \Omega_K = \text{TopK}(A, K)
\end{equation}
where \( \Omega_K \) denotes the set of selected dominant frequencies. The corresponding spectrum of periodic components is extracted as:
\begin{equation}
X_F^{\text{season}}(\omega) = 
\begin{cases}
X_F(\omega), & \omega \in \Omega_K \\
0, & \text{otherwise},
\end{cases}
\end{equation}
This operation effectively filters out weak noise spectra, retaining only the high-energy structures that are statistically significant. Subsequently, we transform back to the time domain:
\[
X^{\text{season}} = \text{IFFT}(X_F^{\text{season}}),
\]
obtaining sequence segments with prominent periodicity. The remaining part naturally constitutes the trend component:
\begin{equation}
    X^{\text{trend}} = X - X^{\text{season}}
\end{equation}

\begin{figure}[t]
    \centering
    \includegraphics[width=0.8\linewidth]{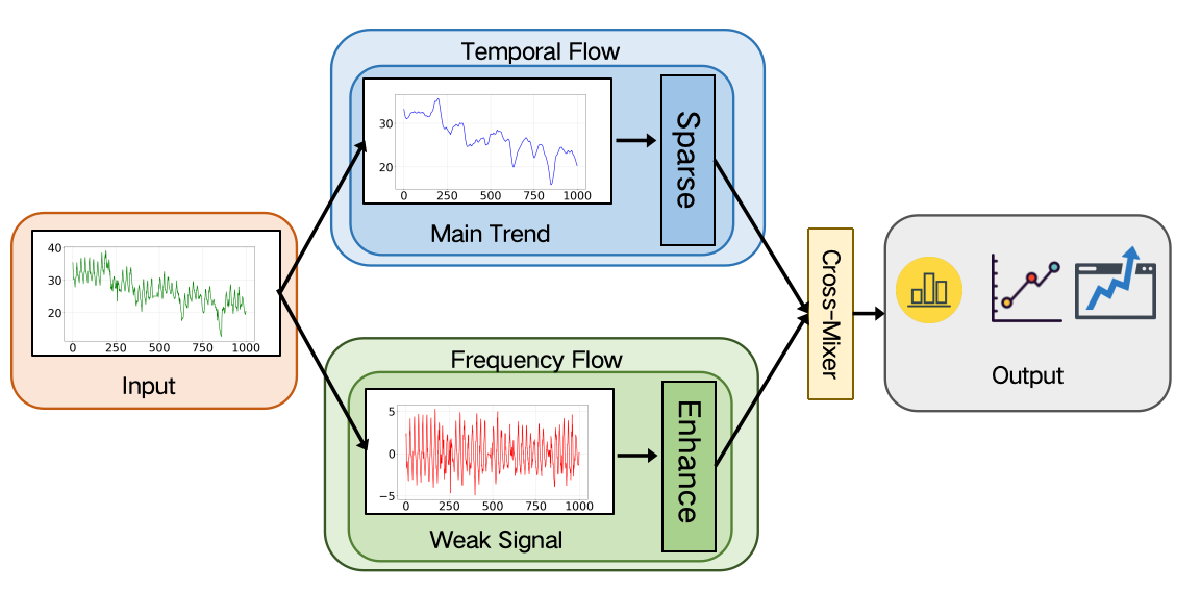}
    \caption{Framework of SDMixer.}
    \label{fig:model}
\end{figure}

Through this learnable spectral truncation mechanism, trend and seasonal information are no longer dependent on manual prior assumptions but instead generate complementary structures in a data-driven manner. SDMixer constructs a sparse temporal domain modeler and a frequency domain enhancement network for in-depth modeling, respectively, and finally fuses the results to obtain the prediction:
\begin{equation}
    \hat{Y} = \text{Fusion}(f_\theta^{\text{T}}(X^{\text{trend}}), f_\theta^{\text{F}}(X^{\text{season}}))
\end{equation}
where \( \hat{Y} \in \mathbb{R}^{B \times L' \times C} \) is the prediction output, and \( L' \) is the prediction sequence length. The overall framework of the model is illustrated in Figure \ref{fig:model}.

\subsection{Sparse Temporal Flow}

Trend information typically manifests as slow changes and long-term dependencies, while multivariate noise can interfere with trend modeling. Therefore, there is a need to explicitly weaken the contributions of irrelevant variables in the time domain. First, linear projection is used to extract potential trend features:
\begin{equation}
    H_T = X^{\text{trend}} W_T
\end{equation}
where \( W_T \in \mathbb{R}^{C \times C} \) is a learned transformation for variable interactions. To adaptively highlight truly effective variable dependencies at each time step, we introduce a magnitude-based sparse gating function:
\begin{equation}
    G_T = \text{SparseTopK}(H_T)
\end{equation}
where only the top-\( k = \lceil \alpha C \rceil \) channels with the largest magnitudes at each time step are retained, and the remaining contributions are dynamically masked as noise or weak dependencies. Furthermore, the sparsity-enhanced trend representation is obtained as:
\begin{equation}
    Z_T = H_T \odot G_T
\end{equation}
where \( \odot \) denotes the element-wise multiplication operation. Subsequently, temporal mixing is performed via an MLP to learn long-term smooth structures:
\begin{equation}
    F_T = \text{MLP}_T(Z_T)
\end{equation}
This design ensures that trend modeling relies only on key variables, making its expression more stable and robust to noise.

\subsection{Frequency Flow}

Seasonal sequences often exhibit characteristics of multi-period superposition, whose patterns are difficult to capture directly in the time domain. Therefore, a frequency-domain representation is adopted for \( X^{\text{season}} \):
\[
X_F^{\text{season}} = \text{FFT}(X^{\text{season}}).
\]
Given that weak periodic signals may have small magnitudes but contain key dynamic information, we enhance the expressive power of their real parts:
\begin{equation}
    Z_F = \text{Enhance}(\Re(X_F^{\text{season}})) + i \cdot \Im(X_F^{\text{season}})
\end{equation}
where \( \text{Enhance}(\cdot) \) is a trainable linear module designed to amplify periodic features that were overshadowed by trends. To return to an interpretable temporal space, we perform an inverse transform:
\[
F_F = \text{IFFT}(Z_F),
\]
endowing it with temporal alignability to facilitate fusion, while retaining the advantages of frequency-domain modeling for capturing long-term dependencies.

\subsection{Sparse Cross-Mixer}

The influence of trends and periods on the future is dynamically variable, thus requiring an adaptive fusion mechanism to adjust their dominance. We project to generate query, key, and value vectors respectively:
\[
Q = F_T W_Q, \quad K = F_F W_K, \quad V = F_F W_V,
\]
where \( W_Q, W_K, W_V \) are trainable parameters. The query is derived from the trend representation, implying that the fusion is domained by the trend to selectively incorporate complementary information. Attention weights are obtained through sparse selection:
\begin{equation}
    \alpha = \text{TopK}
\left(
\text{Softmax}\left(
\frac{QK^\top}{\sqrt{C}}
\right)
\right),
\end{equation}
where \( \alpha \in \mathbb{R}^{B \times L \times L} \) retains only the periodic dependencies that are highly relevant to the trend. The final fused representation is:
\begin{equation}
    H = F_T + \sigma(\gamma) (\alpha V)
\end{equation}
where \( \gamma \) is a learnable scaling factor and \( \sigma(\cdot) \) is the Sigmoid function, used to control the weight of frequency-domain contributions, enabling the fused result to be compatible with both stationary (trend) and oscillatory (periodic) patterns.

\section{Experiments and Results}
\subsection{Experimental Setup}
\subsubsection{Datasets}
This study conducts evaluations on multiple public multivariate time series forecasting benchmark datasets, including multi-source signals from real-world application scenarios such as meteorology, transportation, and energy. All data are normalized using mean-variance normalization, and detailed information about the datasets is presented in Table \ref{tab:data}.

\begin{table}[t]
\centering
\caption{Dataset detailed descriptions. The dataset size is organized in (Train, Validation, Test). Forecastability is calculated by one minus the entropy of Fourier decomposition of time series \cite{forecastability}; a larger value indicates better predictability.}
\label{tab:data}
\begin{tabular}{lccccc}
\toprule
Dataset     & Dim  & Dataset Size           & Forecastability* & Information & $\frac{\text{cov(Season)}}{\text{cov(Trend)}}$ \\
\midrule
ETTm1       & 7    & $(34465, 11521, 11521)$ & 0.46             & Temperature & 0.073992 \\
ETTm2       & 7    & $(34465, 11521, 11521)$ & 0.55             & Temperature & 0.039469 \\
ETTh1       & 7    & $(8545, 2881, 2881)$    & 0.38             & Temperature & 0.490909 \\
ETTh2       & 7    & $(8545, 2881, 2881)$    & 0.45             & Temperature & 0.136543 \\
Electricity & 321  & $(18317, 2633, 5261)$   & 0.77             & Electricity & 11.836820 \\
Exchange    & 8    & $(5312, 760, 1516)$     & 0.42             & Economy     & 0.001885 \\
Weather     & 21   & $(36792, 5271, 10540)$  & 0.75             & Weather     & 0.005176 \\
\bottomrule
\end{tabular}
\end{table}

\subsubsection{Parameter Settings}
Experiments are performed on two NVIDIA GeForce RTX 2080 Ti GPUs (12GB each). The lookback window is uniformly set to 96, while the prediction lengths are configured as 96, 192, 336, and 720. Hyperparameters such as batch size, number of epochs, and learning rate are optimized on the validation set. Mean Squared Error (MSE) is adopted as the loss function, and Adam is used as the optimizer.

\subsubsection{Baseline Models}
To comprehensively verify the performance advantages of SDMixer, we select representative models in the field of multivariate time series forecasting as baselines, covering attention-based structures (iTransformer\cite{iTransformer}, PatchTST\cite{PatchTST}), lightweight linear structures (DLinear\cite{DLinear}), decomposition-based models (TimesNet\cite{TimesNet}, Autoformer\cite{Autoformer}, TimeMixer\cite{TimeMixer}), and frequency-domain processing-based models (WPMixer\cite{WPMixer}, FEDformer\cite{FEDformer}). Except for WPMixer, the results of other models are obtained from \cite{TimeMixer++}. Since the original paper of WPMixer did not disclose results with a lookback window of 96, we reran and tuned WPMixer to avoid underestimating the baseline performance.

\begin{table}[]
\begin{tabular}{cccccccccccc}
\noalign{\hrule height 1pt} % 顶部粗线
                             &     & \multicolumn{2}{c}{SDMixer}     & \multicolumn{2}{c}{WPMixer}     & \multicolumn{2}{c}{TimeMixer}   & \multicolumn{2}{c}{iTransformer} & \multicolumn{2}{c}{PatchTST}    \\
                             &     & MSE            & MAE            & MSE            & MAE            & MSE            & MAE            & MSE             & MAE            & MSE            & MAE            \\ \noalign{\hrule height 1pt} % 顶部粗线
\multirow{5}{*}{Weather}     & 96  & 0.168          & 0.221          & \textbf{0.163} & \textbf{0.205} & 0.163          & 0.209          & 0.174           & 0.214          & 0.186          & 0.227          \\
                             & 192 & 0.210          & 0.256          & 0.208          & \textbf{0.245} & \textbf{0.208} & 0.250          & 0.221           & 0.254          & 0.234          & 0.265          \\
                             & 336 & \textbf{0.256}          & 0.291          & 0.263          & \textbf{0.287} & 0.259 & 0.287          & 0.278           & 0.296          & 0.284          & 0.301          \\
                             & 720 & \textbf{0.319} & \textbf{0.336} & 0.339          & 0.339          & 0.339          & 0.341          & 0.358           & 0.347          & 0.356          & 0.349          \\
                             & AVG & \textbf{0.238} & 0.276          & 0.243          & \textbf{0.269} & 0.240          & 0.271          & 0.258           & 0.278          & 0.265          & 0.285          \\ \hline
\multirow{5}{*}{Electricity} & 96  & \textbf{0.137} & \textbf{0.234} & 0.150          & 0.241          & 0.153          & 0.247          & 0.148           & 0.240          & 0.190          & 0.296          \\
                             & 192 & \textbf{0.152} & \textbf{0.249} & 0.163          & 0.253          & 0.166          & 0.256          & 0.162           & 0.253          & 0.199          & 0.304          \\
                             & 336 & \textbf{0.169} & \textbf{0.266} & 0.180          & 0.270          & 0.185          & 0.277          & 0.178           & 0.269          & 0.217          & 0.319          \\
                             & 720 & \textbf{0.209} & \textbf{0.300} & 0.219          & 0.305          & 0.225          & 0.310          & 0.225           & 0.317          & 0.258          & 0.352          \\
                             & AVG & \textbf{0.167} & \textbf{0.262} & 0.178          & 0.267          & 0.182          & 0.272          & 0.178           & 0.270          & 0.216          & 0.318          \\ \hline
\multirow{5}{*}{Exchange}    & 96  & \textbf{0.082} & \textbf{0.200} & 0.093          & 0.219          & 0.090          & 0.235          & 0.086           & 0.206          & 0.088          & 0.205          \\
                             & 192 & \textbf{0.174} & \textbf{0.296} & 0.199          & 0.317          & 0.187          & 0.343          & 0.177           & 0.299          & 0.176          & 0.299          \\
                             & 336 & 0.322          & 0.410          & 0.378          & 0.440          & 0.353          & 0.473          & 0.331           & 0.417          & \textbf{0.301} & \textbf{0.397} \\
                             & 720 & \textbf{0.840}          & \textbf{0.691} & 2.030          & 0.978          & 0.934          & 0.761          & 0.847           & 0.691          & 0.901          & 0.714          \\
                             & AVG & \textbf{0.355}          & \textbf{0.399} & 0.675          & 0.489          & 0.391          & 0.453          & 0.360           & 0.403          & 0.367          & 0.404          \\ \hline
\multirow{5}{*}{ETTh1}       & 96  & \textbf{0.371}          & 0.398          & 0.371 & \textbf{0.393} & 0.375          & 0.400          & 0.386           & 0.405          & 0.460          & 0.447          \\
                             & 192 & \textbf{0.413} & 0.426          & 0.425          & \textbf{0.417} & 0.429          & 0.421          & 0.441           & 0.512          & 0.477          & 0.429          \\
                             & 336 & \textbf{0.443} & 0.444          & 0.465          & \textbf{0.434} & 0.484          & 0.458          & 0.487           & 0.458          & 0.546          & 0.496          \\
                             & 720 & \textbf{0.448}          & \textbf{0.485}          & 0.453 & 0.448 & 0.498          & 0.482          & 0.503           & 0.491          & 0.544          & 0.517          \\
                             & AVG & \textbf{0.416} & 0.438          & 0.428          & \textbf{0.423} & 0.447          & 0.440          & 0.454           & 0.447          & 0.516          & 0.484          \\ \hline
\multirow{5}{*}{ETTh2}       & 96  & \textbf{0.269} & \textbf{0.335} & 0.285          & 0.337          & 0.289          & 0.341          & 0.297           & 0.349          & 0.308          & 0.355          \\
                             & 192 & \textbf{0.333} & \textbf{0.379} & 0.364          & 0.392          & 0.372          & 0.392          & 0.380           & 0.400          & 0.393          & 0.405          \\
                             & 336 & \textbf{0.360} & \textbf{0.406} & 0.378          & 0.407          & 0.386          & 0.414          & 0.428           & 0.432          & 0.427          & 0.436          \\
                             & 720 & 0.412          & 0.445          & 0.418          & 0.436          & \textbf{0.412} & \textbf{0.434} & 0.427           & 0.445          & 0.436          & 0.450          \\
                             & AVG & \textbf{0.343} & \textbf{0.391} & 0.361          & 0.393          & 0.364          & 0.395          & 0.383           & 0.407          & 0.391          & 0.411          \\ \hline
\multirow{5}{*}{ETTm1}       & 96  & 0.317          & 0.359          & \textbf{0.316} & \textbf{0.351} & 0.320          & 0.357          & 0.334           & 0.368          & 0.352          & 0.374          \\
                             & 192 & \textbf{0.335} & \textbf{0.366} & 0.358          & 0.376          & 0.361          & 0.381          & 0.390           & 0.393          & 0.374          & 0.387          \\
                             & 336 & \textbf{0.365} & \textbf{0.384} & 0.387          & 0.397          & 0.390          & 0.404          & 0.426           & 0.420          & 0.421          & 0.414          \\
                             & 720 & \textbf{0.413} & \textbf{0.410} & 0.445          & 0.430          & 0.454          & 0.441          & 0.491           & 0.459          & 0.462          & 0.449          \\
                             & AVG & \textbf{0.357} & \textbf{0.379} & 0.377          & 0.389          & 0.381          & 0.395          & 0.407           & 0.410          & 0.406          & 0.407          \\ \hline
\multirow{5}{*}{ETTm2}       & 96  & \textbf{0.161} & \textbf{0.252} & 0.171          & 0.252          & 0.175          & 0.258          & 0.180           & 0.264          & 0.183          & 0.270          \\
                             & 192 & \textbf{0.216} & \textbf{0.291} & 0.234          & 0.295          & 0.237          & 0.299          & 0.250           & 0.309          & 0.255          & 0.314          \\
                             & 336 & \textbf{0.268} & \textbf{0.327} & 0.290          & 0.333          & 0.298          & 0.340          & 0.311           & 0.348          & 0.309          & 0.347          \\
                             & 720 & \textbf{0.351} & \textbf{0.384} & 0.387          & 0.390          & 0.391          & 0.396          & 0.412           & 0.407          & 0.412          & 0.404          \\
                             & AVG & \textbf{0.249} & \textbf{0.313} & 0.270          & 0.317          & 0.275          & 0.323          & 0.288           & 0.332          & 0.290          & 0.334          \\ \noalign{\hrule height 1pt} % 顶部粗线
\end{tabular}
\caption{Comparative experimental results with typical baselines. The best performance is in bold.}
\label{tab:cmp}
\end{table}

\begin{table}[t]
\centering
\caption{Comparative experimental results with more baselines. For each dataset, the result is the average over the four prediction lengths (96, 192, 336, 720). The best performance is in bold.}
\label{tab:cmp2}
\begin{tabular}{lcccccccccccc}
\toprule
\multirow{2}{*}{Dataset} & \multicolumn{2}{c}{SDMixer} & \multicolumn{2}{c}{TimesNet} & \multicolumn{2}{c}{DLinear} & \multicolumn{2}{c}{SCINet} & \multicolumn{2}{c}{FEDformer} & \multicolumn{2}{c}{Autoformer} \\
\cmidrule(lr){2-13} % 从第2列画到第13列，并且左右留出空隙
                         & MSE & MAE & MSE & MAE & MSE & MAE & MSE & MAE & MSE & MAE & MSE & MAE \\
\midrule
Weather     & \textbf{0.238} & \textbf{0.276} & 0.259 & 0.287 & 0.265 & 0.315 & 0.292 & 0.363 & 0.309 & 0.360 & 0.338 & 0.382 \\
Electricity & \textbf{0.167} & \textbf{0.262} & 0.192 & 0.304 & 0.225 & 0.319 & 0.268 & 0.365 & 0.214 & 0.327 & 0.227 & 0.338 \\
Exchange    & \textbf{0.355} & \textbf{0.399} & 0.416 & 0.443 & 0.354 & 0.414 & 0.750 & 0.626 & 0.519 & 0.429 & 0.613 & 0.539 \\
ETTh1       & \textbf{0.426} & \textbf{0.438} & 0.458 & 0.450 & 0.461 & 0.457 & 0.747 & 0.647 & 0.498 & 0.484 & 0.496 & 0.487 \\
ETTh2       & \textbf{0.343} & \textbf{0.391} & 0.414 & 0.427 & 0.563 & 0.519 & 0.954 & 0.723 & 0.437 & 0.449 & 0.450 & 0.459 \\
ETTm1       & \textbf{0.357} & \textbf{0.379} & 0.400 & 0.406 & 0.404 & 0.408 & 0.485 & 0.481 & 0.448 & 0.452 & 0.588 & 0.517 \\
ETTm2       & \textbf{0.249} & \textbf{0.313} & 0.291 & 0.333 & 0.354 & 0.402 & 0.954 & 0.723 & 0.305 & 0.349 & 0.327 & 0.371 \\
\bottomrule
\end{tabular}
\end{table}

\subsection{Comparative Experimental Results and Analysis}
The comparative experimental results of SDMixer with typical state-of-the-art models  are shown in Table \ref{tab:cmp}, and the results compared with other models are presented in Table \ref{tab:cmp2}. Comparative experiments demonstrate that SDMixer achieves optimal or suboptimal error performance under different prediction lengths and multi-data distribution conditions, with particularly significant advantages in long-term forecasting and high-noise data. Through the collaborative modeling of the dual-stream structure, SDMixer can effectively overcome the learning bias of existing models towards high-energy dominant signals, ensuring that weak but critical periodic information is no longer overwhelmed during inference. Thus, it achieves the unification of improved trend stability and fine-grained feature recovery capability.

\subsection{Ablation Experiments}
To investigate the necessity of each component in the model structure, we conduct ablation analysis by removing the sparse temporal flow, frequency-domain weak signal enhancement flow, and sparse cross-mixer module one by one. Detailed results are shown in Table \ref{tab:amb}. Experiments reveal that removing any module leads to a significant degradation in prediction performance, and the full model outperforms all ablated versions in terms of MSE and MAE. This trend clearly demonstrates that the three core modules in the SDMixer architecture all make substantive contributions, collectively forming an important mechanism to resist structural bias.

\subsection{Further Research}
\begin{figure}[t]
    \centering
    \includegraphics[width=0.7\linewidth]{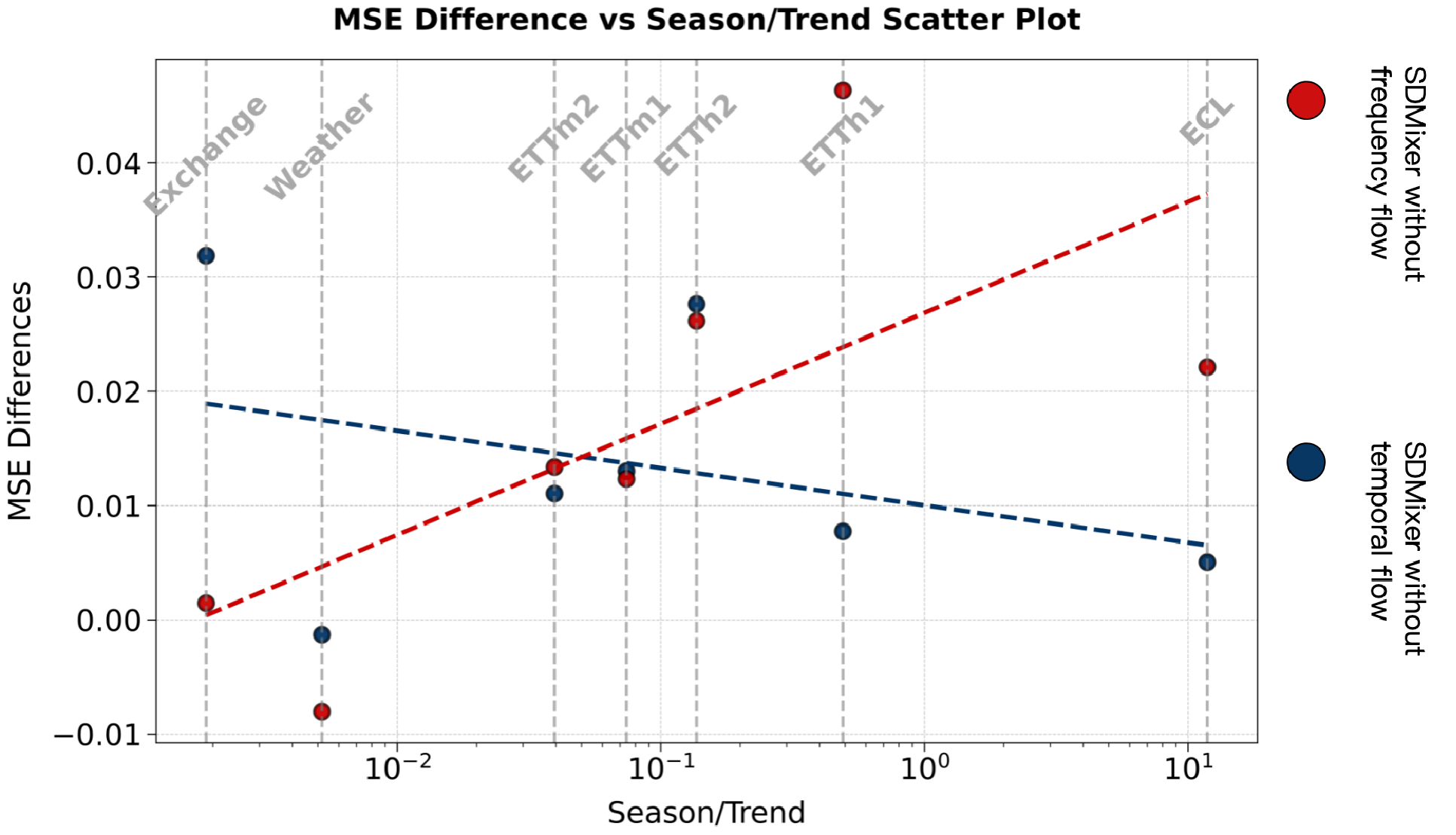}
    \caption{Performance of ablation models on different datasets (errors for prediction length 96).}
    \label{fig:amb}
\end{figure}

To further explore the impact of trend and periodic components on different datasets, we analyze the results of ablation experiments across various data. First, all datasets are normalized using mean-variance normalization for consistency. Then, the data are decomposed into trend and seasonal components using moving average. Subsequently, we calculate the ratio of the covariance of the seasonal component to that of the trend component, which reflects the contribution of each component to the data composition. The results are presented in Table \ref{tab:data}. We compute the MSE errors between the results of SDMixer without the frequency component, SDMixer without the temporal component, and the full model, respectively. Scatter plots are drawn to observe the relationship between the error values and the contribution ratio of the two components, with results shown in Figure \ref{fig:amb}. As illustrated in the figure, the ratio of the covariance of the seasonal component to that of the trend component exhibits a negative correlation with the error of SDMixer without the temporal component, while showing a positive correlation with the error of SDMixer without the frequency component. This further confirms the role of each module in enhancing data feature representation.

\begin{table}[t]
\centering
\caption{Ablation experiment results. For each dataset, the result is the average over the four prediction lengths (96, 192, 336, 720). The best performance is in bold.}
\label{tab:amb}
\begin{tabular}{lcccccccccc}
\toprule
\multirow{2}{*}{Model Variant} & \multirow{2}{*}{Metric} & \multicolumn{7}{c}{Datasets} & \multirow{2}{*}{AVG} \\
\cmidrule(lr){3-9} % 只画第3到9列的横线，并且左右留出空隙
                              &                          & ECL   & ETTh1 & ETTh2 & ETTm1 & ETTm2 & Exchange & Weather \\
\midrule
\multirow{2}{*}{No Cross}     & MSE                     & 0.170 & 0.436 & 0.364 & 0.381 & 0.278 & 0.375    & 0.279   & 0.326 \\
                              & MAE                     & 0.271 & 0.448 & 0.421 & 0.384 & 0.359 & 0.429    & 0.338   & 0.379 \\
\midrule
\multirow{2}{*}{No Time}      & MSE                     & 0.178 & 0.424 & 0.344 & 0.356 & 0.268 & 0.408    & 0.249   & 0.318 \\
                              & MAE                     & 0.279 & 0.440 & 0.395 & 0.383 & 0.329 & 0.439    & 0.287   & 0.365 \\
\midrule
\multirow{2}{*}{No Freq}      & MSE                     & 0.193 & 0.469 & 0.367 & 0.374 & 0.258 & 0.355    & 0.239   & 0.322 \\
                              & MAE                     & 0.301 & 0.473 & 0.410 & 0.395 & 0.320 & 0.400    & 0.278   & 0.368 \\
\midrule
\multirow{2}{*}{Full}         & MSE                     & \textbf{0.167} & \textbf{0.426} & \textbf{0.343} & \textbf{0.357} & \textbf{0.249} & \textbf{0.355}    & \textbf{0.238}   & \textbf{0.305} \\
                              & MAE                     & \textbf{0.262} & \textbf{0.438} & \textbf{0.391} & \textbf{0.379} & \textbf{0.313} & \textbf{0.399}    & \textbf{0.276}   & \textbf{0.351} \\
\bottomrule
\end{tabular}
\end{table}

\section{Conclusion}
To tackle multi-scale characteristics and weak signal obscuration in multivariate time series forecasting, this paper proposes SDMixer, it enhances weak periodic signals and filters invalid dependencies, balances efficiency and performance via lightweight mixers, and achieves adaptive feature fusion. Experiments show SDMixer outperforms mainstream models on diverse real-world datasets and across prediction lengths, offering a new approach for time series forecasting.

\section{AI Usage Statement}
This paper used Doubao for translation and polishing. We have carefully checked the translated content to ensure its accuracy.

\bibliographystyle{splncs04}
\bibliography{ref.bib}
\end{document}